\documentclass{article}

\usepackage{arxiv}

\usepackage[utf8]{inputenc} 
\usepackage[T1]{fontenc}    
\usepackage{hyperref}       
\usepackage{url}            
\usepackage{booktabs}       
\usepackage{nicefrac}       
\usepackage{microtype}      
\usepackage{lipsum}
\usepackage{graphicx}
\graphicspath{ {./images/} }

\usepackage{cite}
\usepackage{amsmath,amssymb,amsfonts}
\usepackage{algorithmic}
\usepackage{textcomp}
\usepackage{xcolor}
\usepackage{multirow}

\title{Is Sliding Window All You Need? An Open Framework for Long-Sequence Recommendation}

\author{
 Sayak Chakrabarty$^\dagger$ \\
  Department of Computer Science\\
  Northwestern University\\
  Evanston, IL 60208, USA \\
  \texttt{sayakchakrabarty2025@u.northwestern.edu}
   \And
 Souradip Pal$^\dagger$ \\
  School of Electrical and Computer Engineering\\
  Purdue University\\
  West Lafayette, IN 47906, USA \\
  \texttt{pal43@purdue.edu} \\
}

\begin{document}
\maketitle

\def\thefootnote{\dag}\footnotetext{These authors contributed equally to this work.}
\def\thefootnote{\arabic{footnote}}

\begin{abstract}
Long interaction histories are central to modern recommender systems, yet training with long sequences is often dismissed as impractical under realistic memory and latency budgets. This work demonstrates that it is not only practical but also effective—\emph{at academic scale}. We release a complete, end-to-end framework that implements industrial-style long-sequence training with sliding windows, including all data processing, training, and evaluation scripts. Beyond reproducing prior gains, we contribute two capabilities missing from earlier reports: (i) a \textbf{runtime-aware ablation study} that quantifies the accuracy–compute frontier across windowing regimes and strides, and (ii) a \textbf{novel k-shift embedding} layer that enables million-scale vocabularies on commodity GPUs with negligible accuracy loss. Our implementation trains reliably on modest university clusters while delivering competitive retrieval quality (e.g., up to \(+\!6.04\%\) MRR and \(+\!6.34\%\) Recall@10 on Retailrocket) with $\sim 4\times$ training-time overheads. By packaging a robust pipeline, reporting training time costs, and introducing an embedding mechanism tailored for low-resource settings, we transform long-sequence training from a closed, industrial technique into a practical, open, and extensible methodology for the community.
\end{abstract}


\section{Introduction}

Long-range user behavior is a rich signal for recommendation quality, but bringing long histories into training often collides with practical constraints: fixed inference budgets, tight memory envelopes, and limited accelerator availability. Sliding-window training is a simple and powerful idea—expose broad temporal coverage over epochs while honoring a short per-step context—but existing descriptions have been difficult to adopt: code is unavailable, runtime trade-offs are under-reported, and scaling to million-scale vocabularies typically assumes industrial hardware.

This paper closes those gaps with a complete, \emph{ready-to-run} pipeline and several contributions that make long-sequence training straightforward and cost-conscious on academic infrastructure:

\begin{enumerate}
    \item \textbf{Open Implementation.} Our work provides a full training stack—data preparation, sliding-window approach, evaluation, logging, and configurations - so researchers can run, modify, and extend long-sequence experiments without reverse-engineering.
    \item \textbf{Runtime-Aware Ablations.} Our study shows the missing dimension in prior reports: a systematic accuracy–compute analysis across windowing modes (All-Sliding, Mixed), strides, and model sizes, with training time recorded end-to-end. These results expose actionable speed–quality frontiers rather than single-point metrics.
    \item \textbf{Novel \emph{k}-Shift Embeddings.} We introduce a compact embedding layer that scales to million-item vocabularies on commodity GPUs by using multiple seed-shifted lookups into a shared table with near-collision-free indexing. The design preserves recommendation quality while reducing memory pressure and enabling larger effective vocabularies.
    \item \textbf{Low-Resource Feasibility.} All components run robustly on common university clusters. We show that long-sequence training using \emph{k}-shift embeddings and sliding-window regimes become practical and repeatable for academic labs with this framework.
\end{enumerate}

Using public benchmarks (e.g., Retailrocket), we quantify both retrieval gains and training costs, observe consistent improvements from long-range exposure, and detail where stride and window choices move the compute–quality frontier. The outcome is a transparent blueprint that upgrades sliding-window training from a promising concept to a deployable, efficient practice in open settings.

\section{Related Work}
\begin{figure}[ht]
\centering
  \includegraphics[width=0.6\linewidth]{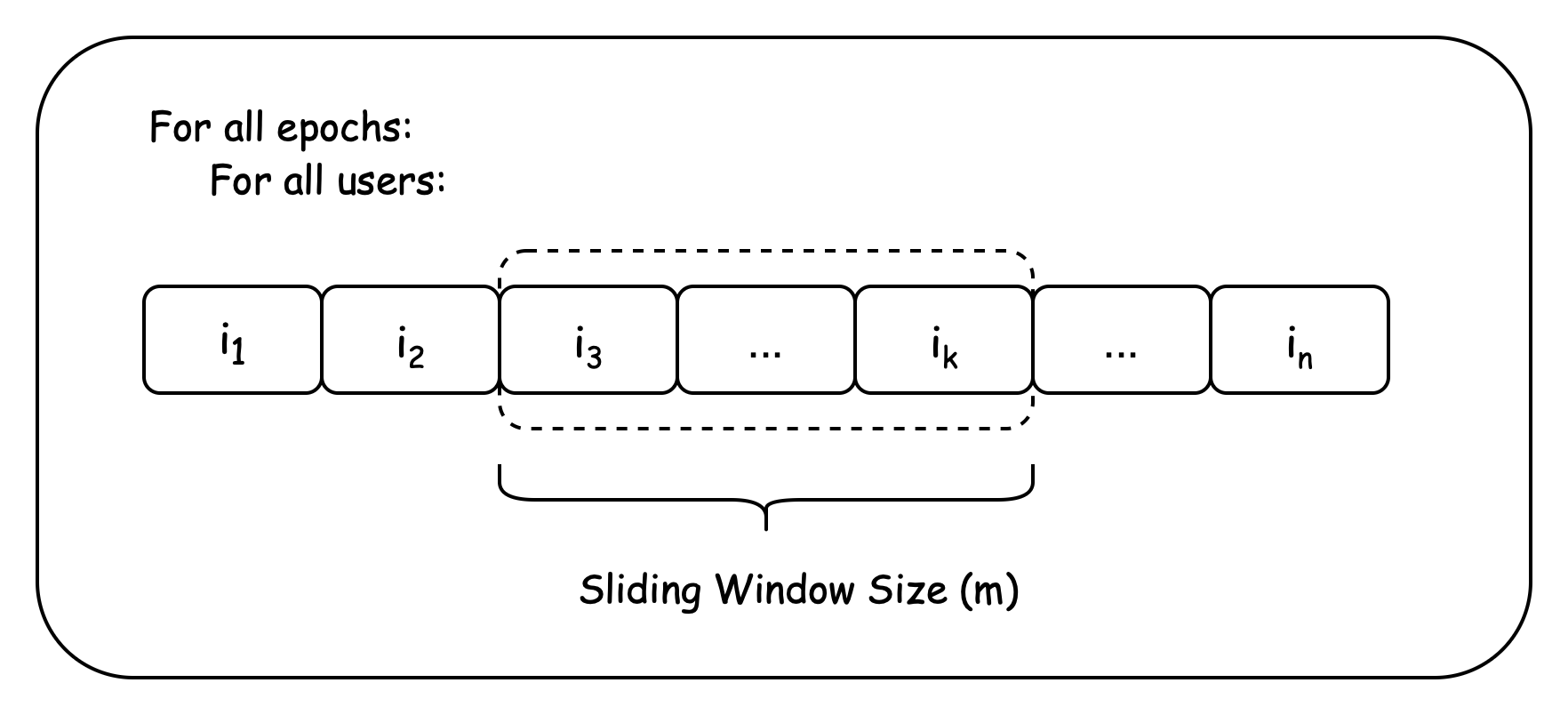}
  \caption{Sliding window training loop}
  \label{fig:training}
\end{figure}

Maximizing the interaction history of users is actively being explored in large RecSys foundation models \cite{geng2023recommendation} to make learning more efficient \cite{zhu2024pose}. \cite{zhai2024actions,datta2023consistency} focuses on a setting with temporally repetitive user behaviors, enabling the use of artificial sparsification to reduce down long sequences. While \cite{joshi2024sliding} focuses on a novel sliding window approach to reduce the context window during training as shown in Figure \ref{fig:training}. It also focuses on data augmentation to improve the richness of representations under inference latency and model capacity constraints. Generally, across large foundation models, there has been work on extending the context windows to a very large token size without affecting the model performance \cite{ding2024longrope}. Even research on context windows for large language and vision models \cite{makarychev2023single,chakrabarty2024free,zhang2023sockdef,Chakrabarty2025,bolonkin2024judicial,chakrabarty2025timeconstrainedrecsys,chakrabarty2025mm,li2023gpt4rec,chakrabarty2026pixrec} employed for sequential recommendation tasks on various public datasets has gained popularity in recent times but those approaches are largely orthogonal to the general goal of exploiting long sequence corpora without increasing model context windows.

\section{Motivation and Approach}
The key motivation behind this word lies in ensuring transparent, replicable, and extensible recommender-system research for long-range behavioral context in academia. Although the study \cite{joshi2024sliding} provides a high-level algorithmic description of the sliding window training technique and performance metrics on a large interaction dataset, we encountered several \textit{research questions}(\textbf{RQ}) and had to make various assumptions along the way. Given these uncertainties, our approach was as follows:
\begin{enumerate}
    \item \textbf{Assumption and Implementation}: For replicating the results, we select a simple auto-regressive recommendation model and implement the sliding window method based on the assumption that the method is model agnostic.
    \item \textbf{Experimental Validation}: Because the exact dataset from paper \cite{joshi2024sliding} is not publicly available, we used a similar publicly accessible dataset for our validation. This dataset follows a similar user interaction pattern to the one used in the original experiments thus making them ideal for replication.
\end{enumerate}

Based on our approach, we intend to find the answers to the following research question: 
\begin{itemize}
    \item \textbf{RQ1:} How do the sliding-window approach and stride choices trade off retrieval quality versus training-time cost?
    \item \textbf{RQ2:} Can our pipeline deliver consistent long-sequence gains on commodity GPU?
    \item \textbf{RQ3:} Do \textit{k}-shift embeddings maintain recommendation quality while materially reducing memory footprint for large vocabularies?
\end{itemize}

\section{Experimental Setup}
\subsection{Datasets}
The paper \cite{joshi2024sliding} used a large user interactions dataset of the order of 250M users and their interactions with the items in the content library. Such interaction sequences of users might include video plays, video likes, adding to watch list, opening video details page, etc. These interactions could span over long periods ranging from weeks to months. Since we do not have access to this non-disclosed dataset, we leverage the following publicly available datasets, as shown in Table \ref{table:datasets}, that have similar characteristics. Note that the Retailrocket dataset has the following three types of events, namely \textbf{view}, \textbf{addtocart} or \textbf{transaction}. These behaviors parallel the interaction types in the dataset used in the original paper, making these two public datasets (Retailrocket and Taobao) closest to it in terms of the sequential recommendation format. These datasets include different types of user interactions which is crucial for the experiments performed, while other recommendation datasets (e.g. MovieLens and Amazon Reviews), even though are of similar size to the original proprietary dataset used by Netflix, are mostly rating datasets that don't include user interactions like clicks, buys, etc. The paper from Netflix by \cite{joshi2024sliding} has used only a single dataset, which doesn't show generalization. Our experiment was conducted on at least two datasets, which shows some form of generalization of the sliding window training approach.

\begin{table}[ht]
\centering
\scriptsize
\caption{Dataset Details}
\label{table:datasets}
\begin{tabular}{lcccc}
\toprule
\textbf{Dataset}  & \textbf{Users} & \textbf{Items} & \textbf{Interactions} & \textbf{Behavior Types} \\
\midrule
\textbf{Taobao}     & 987,994         & 4,162,024  & 100,150,807         &  4      \\
\textbf{Retailrocket}     & 1,407,580         & 417,053         & 2,756,101        & 3   \\
\bottomrule
\end{tabular}
\end{table}

\begin{table}[ht]
\centering
\caption{Default hyperparameters used in the foundation model for Sliding Window Training with Retailrocket dataset}
\label{table:hyperparams}
\begin{tabular}{cc}
    \toprule
    \textbf{Hyperparameter} & \textbf{Default Value} \\
    \midrule
    \texttt{emb\_dim} & 32 \\
    \texttt{n\_layers} & 2 \\
    \texttt{n\_heads} & 4 \\
    \texttt{dropout} & 0.1 \\
    \texttt{window\_size} & 100 \\
    \texttt{learning\_rate} & 0.001 \\
    \bottomrule
\end{tabular}
\end{table}

\begin{figure}[ht]
\centering
  \includegraphics[width=0.8\linewidth]{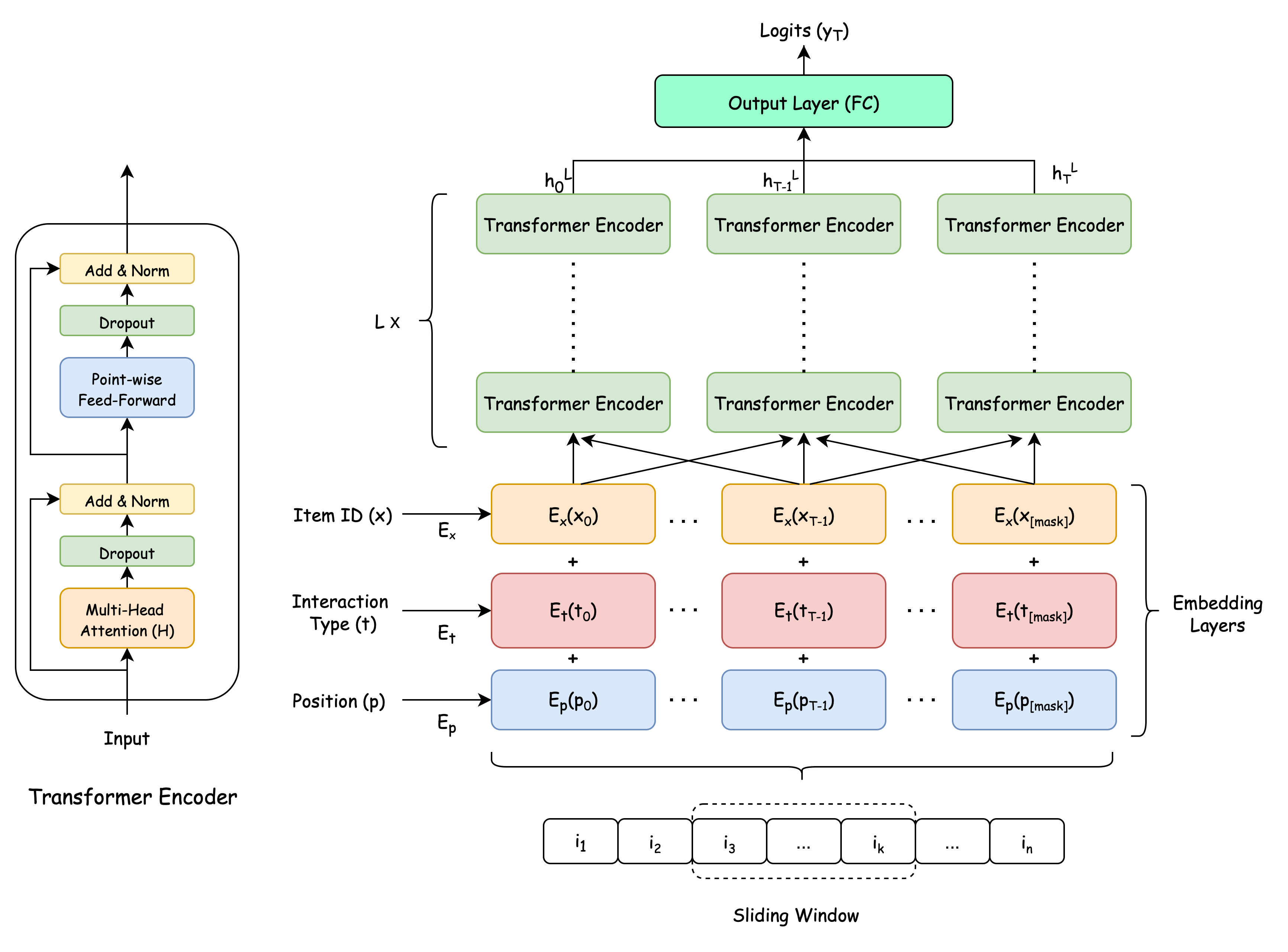}
  \caption{Overview of the RecSys Foundation model architecture used for Sliding Window Training, designed similar to BERT4Rec architecture with the additional embedding layer($E_t$) encoding categorical interaction types($t$) like View, AddToCart \& Transaction}
  \label{fig:model}
\end{figure}

\subsection{Training \& Evaluation}
 \paragraph{\textbf{Model}} Here, a generic encoder-based sequential recommendation model has been used for our experiments that incorporate item embeddings, interaction types, and positional information into the input representation. The model is designed to predict the next relevant item in the sequence based on prior interactions and their types. We use an item ID based representation learning in our foundation model similar to \cite{10.1145/3357384.3357895} but perform training in an auto-regressive fashion on a sequential recommendation dataset with the assumption that the target interaction depends on the previous interactions. However, it is to be noted that the sliding window training technique is assumed to be model agnostic, generalizable to any foundation model. The model architecture consists of the following as shown in Fig. \ref{fig:model}:

\begin{itemize}
    \item \textbf{Item Embedding Layer} ($E_x$): Embedding layer that maps item IDs to a dense vector space of dimension $d$.
    \item \textbf{Interaction Type Embedding Layer} ($E_t$): Embedding layer that encodes the type of interaction (e.g., click, add-to-cart, purchase, etc.) into the same $d$-dimensional space.
    \item \textbf{Position Embedding Layer} ($E_p$): Encodes the positional index of each item in the sequence using learnable embeddings of size $d$.
    \item \textbf{Transformer Encoder} ($G$): A stack of $L$ Transformer encoder layers with multi-head self-attention ($H$ heads) and feedforward sub-layers. A causal mask($m$) is applied to ensure auto-regressive learning.
    \item \textbf{Output Layer} (\texttt{FC}): A linear projection from the Transformer output to the item vocabulary size, used for next-item prediction acting as the classification head.
\end{itemize}

Given an input sequence $x \in \mathbb{Z}^{B \times T}$ of item IDs and corresponding interaction types $t \in \mathbb{Z}^{b \times T}$, the logits $\hat{y}$ are computed as follows:
\begin{align*}
    p_{i} = i, \quad i = 0, \dots, T-1 \\
    e = E_x(x)+E_t(t)+E_p(p) \\
    h = G(e, m)\\
    \hat{y} = \texttt{FC}(h)
\end{align*}
where $B$ is the number of unique item IDs, $b$ is the number of interaction types, $T$ is the maximum sequence length and $p_i$ corresponds to the position of a token in the sequence. 

\paragraph{\textbf{Training}} A potential threat to validity is that the control baseline uses the same default hyperparameters as the sliding-window variants, rather than being independently tuned to optimality. We chose this design intentionally to isolate the effect of the training regime while keeping the model class and optimization setup fixed across comparisons. This makes the comparisons controlled, but it does not rule out the possibility that a more extensively tuned control could narrow part of the gap. We therefore interpret our results as strong evidence that sliding-window training is effective under a matched configuration, while leaving exhaustive baseline tuning to future work. Due to limited clarity on any systematic tuning performed in the original paper and the specifications used for the experimentation, we stick to straight-forward training of a RecSys foundation model from scratch. The hyperparameters of our model training can be found in Table \ref{table:hyperparams}. Using these hyperparameters, 10 epochs of training were performed, with a mixed setup comprising 5 control and 5 sliding (time) runs. 

\paragraph{\textbf{Evaluation}} During evaluation, we calculate the accuracy measures (MRR, Recall and Perplexity) as well as training time, targeting long-sequence training under context limits as per the evaluation methods in the original paper. We compare our results obtained from different sliding variants (All-Sliding, Mixed, All-Sliding + \textit{Strides}) to a naive control approach without any sliding window, that serves as our baseline.



\begin{table*}[ht]
\centering
\caption{Results showing the percentage change in performance for Sliding and Mixed methods relative to Control method}
\label{tab:results}
\begin{tabular}{lcccccc}
\toprule
\textbf{Dataset} & \textbf{Model}  & \textbf{Perplexity} & \textbf{MRR} & \textbf{Recall} & \textbf{Training Time} \\
\midrule
\multirow{4}*{Retailrocket}
& Control     &—         &—         &—        &—    \\
& All-Sliding  &-52.22\%  &+6.04\%   &+6.34\%  &+403.44\%  \\
& Mixed-500    &-43.02\%  &+3.78\%   &+4.56\%  &+188.42\%  \\
& Mixed-1000   &-46.34\%  &+3.36\%  &+4.36\%  &+217.16\%  \\
\bottomrule
\end{tabular}
\end{table*}


\begin{table*}[ht]
\centering
\caption{Ablation study showing the percentage change of strided variants relative to the All-Sliding baseline}
\label{tab:ablation}
\begin{tabular}{lcccc}
\toprule
\textbf{Model} & \textbf{Perplexity} & \textbf{MRR} & \textbf{Recall} & \textbf{Training Time} \\
\midrule
All-Sliding                     &—         &—         &—        &—  \\
All-Sliding + \textit{Stride = 2}             & $+13.4\%$ & $-1.6\%$ & $-1.1\%$ & $-31.8\%$ \\
All-Sliding + \textit{Stride = 4}             & $+25.6\%$ & $-3.3\%$ & $-3.0\%$ & $-51.7\%$ \\
\bottomrule
\end{tabular}
\end{table*}

\section{Results and Discussion}


\subsection{Results} 
As described in \cite{joshi2024sliding}, \textit{All-Sliding} uses a sliding window training loop as in Figure \ref{fig:training} for all $N$ epochs without any focus on recent interactions. \textit{Mixed-500} uses a mixture of fixed recent \& sliding window epochs where sliding windows go as far back as $500$ item interaction events, and \textit{Mixed-1000} where we slide far back as $1000$ events. To compare our reproducibility, we also choose our evaluation metrics as perplexity, Mean Reciprocal Rank \cite{Craswell2009} (MRR) for next item prediction and recall for item embedding quality. Our results align well with the results in \cite{joshi2024sliding} as shown in Table \ref{tab:results}, offering a strong indication that the technique is robust across long contexts.

A notable difference from \cite{joshi2024sliding} is that, on Retailrocket, our best result is obtained by \textit{All-Sliding} rather than \textit{Mixed-1000}. We believe this difference is primarily dataset-dependent. Retailrocket contains shorter and relatively denser user histories, so repeatedly exposing the model to older interactions can still be beneficial because those histories remain semantically coherent within a modest sequence horizon. In contrast, a mixed strategy that prioritizes recent windows may be more advantageous when long histories exhibit stronger temporal drift or when very old interactions become weak predictors of the next item. Put differently, the usefulness of \textit{Mixed-1000} depends on how quickly user intent changes over time and on whether older events remain predictive after accounting for the fixed context size used at training and inference. This suggests that the choice between \textit{All-Sliding} and mixed variants should not be treated as universal; rather, it should be viewed as a function of sequence-length distribution, temporal drift, and interaction sparsity in the target dataset.

\subsection{Further Experiments}
Experiments were also performed on the Taobao \href{https://tianchi.aliyun.com/dataset/649?lang=en-us}{User Behavior dataset} provided by Alibaba \cite{10356156} which contains 4 different types of behaviors: \textbf{pv} (page view of an item's detail page, equivalent to an item click), \textbf{buy} (purchase an item), \textbf{cart} (add an item to the shopping cart) and \textbf{fav} (favor an item). Note that for this dataset, the items with interactions less than a certain threshold (say 50 interactions) were filtered out to reduce the large vocabulary size.

\textbf{Dealing with Large Vocabulary} - Notice that the item embedding layer ($E_x$) depends on the unique item IDs $B$ (vocabulary size) of the dataset. A vocabulary size $B$ of $\sim1M$ (as in the case of the Taobao dataset), can lead to inflated model sizes and infrastructure challenges. To address this, we used K-Shift Embedding \cite{desai2022tradeoffsmodelsizelarge} layers in place of the normal Embedding layers that use a nearly collision-free hashing mechanism allowing different features to share a single embedding table. The K-Shift Embedding approach is a generalization of the QR embeddings \cite{DBLP:journals/corr/abs-1909-02107} using bit-shift operations to simulate multiple pseudo-random accesses into a compact embedding table. This setup supports both standard embedding use and learning ID-to-content embedding mappings directly within the model, reducing the need for external pipelines. The model can be trained efficiently with normalized losses and offers potential for further compression via. reduced-dimensionality or smaller expansion factors. Given a feature value $v$ and feature name $F$, a hashed ID ($z$) is computed using a seeded hash function $hash$:
\[
\text{z} = hash(v, \text{seed}=F) - 2^{63}
\]
The row index for the embedding lookup is computed using bit shifts (similar to bit rotation):
\[
\text{idx}_{i} = \left( \left( z \ll i \right) \mid \left( z \gg (64 - i) \right) \right) \mod B
\]
where $z$ is the hashed ID, $i$ is the shift index (from $0$ to $k-1$) and $B$ is the number of embedding rows. The final embedding is obtained by summing $k$ embeddings and normalizing as follows:
\[
\mathbf{e}(x) = \sum_{i=0}^{k-1} \mathbf{E}[\text{idx}_{i}], \quad \mathbf{e}_{\text{norm}}(x) = \frac{\mathbf{e}(x)}{\|\mathbf{e}(x)\|_2}
\]
Using this approach, we map the filtered list of items to their corresponding embeddings in a much more efficient way as necessary to conduct the experiments in a resource-constrained scenario.\\

The experiments conducted on the Taobao dataset failed to meet expectations due to the substantial vocabulary size and constrained computational resources, which hindered the full reproducibility of the results. The reasons are the following.
\begin{itemize}
    \item Vocabulary Pressure: Retailrocket's 417k items fit in a dense embedding matrix whereas Taobao's $4.1M$ do not. The $k$-shift hash table induces many collisions, blurring semantic distinctions essential for next-item prediction.
    \item Sequence Sparsity: Retailrocket users average 18.2 interactions; Taobao only 9.1. Sliding Window Training benefits scale with the amount of \textit{new} context each window reveals - minimal in Taobao.
    \item Excessive Time Requirement: The implementation of the "All-Sliding" and \textit{Mixed-500} experimental configurations on the Taobao dataset is projected to necessitate a duration that is ten-fold that required for the Retailrocket dataset under equivalent training conditions, rendering it impractical within an academic framework.
    
\end{itemize}

\subsection{Challenges}
Reproducing the industry paper in an academic setting presented several challenges. These are summarized below:

\begin{enumerate}
    \item \textbf{Dataset Construction:} Building both the Retailrocket and Taobao datasets was time-consuming, and the lack of explicit instructions in the original paper required us to make several assumptions.
    
    \item \textbf{Dataset Constraints:} For the Retailrocket dataset, the batch size was reduced from 64 to 32 to prevent CUDA memory errors on an 80GB H100 GPU. Moreover, the embedding size was lowered from 64 to 32 to fit into the available memory.  For the Taobao dataset, a batch size of 16 and an embedding size of 16 were used where 5 training epochs, with 2 control and 3 sliding (time) runs were performed.
    
    \item \textbf{Vocabulary Size:} The large vocabulary (due to too many unique items) in the case of Taobao dataset was addressed by applying hashing with $k$-shift embedding \cite{DBLP:journals/corr/abs-1909-02107}  to reduce the model size. Items with fewer than 50 interactions were filtered out to mitigate data sparsity.
    
    \item \textbf{Hyperparameter Settings:} Since the original paper did not specify hyperparameters, we resorted to using default values throughout.

    \item \textbf{Evaluation Metric:} We report perplexity, MRR, and Recall because they are substantially cheaper to compute than full-ranking metrics such as mAP on large candidate sets, making them more appropriate for a resource-constrained reproducibility study. However, this choice also narrows the scope of our conclusions. In recommendation settings where full-list quality matters more than the rank of the first relevant item, for example, basket construction, complementary-item recommendation, or multi-item discovery metrics such as mAP or NDCG over larger candidate sets may reveal additional trade-offs not captured by MRR and Recall alone. Our findings should therefore be interpreted primarily as evidence about efficient next-item retrieval under constrained evaluation budgets.

\end{enumerate}

\section{Ablation Study}
Since the use of a sliding window increases the time required for training, one interesting study that can be thought of is how to reduce the training time of such models yet use the same approach. One way to do that is to use strides during sliding. Inspired by the employment of strides in learning representations using convolution neural networks, the intriguing question is whether adding strides still provides decent information to capture the long-term interaction pattern. Thus, we do an ablation study where we add strides of 1,2 and 4 on the sliding window method and compare the results. As shown in Table \ref{tab:ablation}, increasing the stride from 2 to 4 leads to a noticeable rise in perplexity and a corresponding drop in performance metrics such as MRR and recall. This outcome aligns with the understanding that skipping more interactions results in less sequential information for the model. However, training time decreases significantly when using a larger stride, making it feasible to train on longer interaction sequences within practical time limits. These observations underscore the trade-offs involved in replicating closed-source industrial studies, where computational efficiency must be weighed against predictive accuracy.

The large perplexity increase at \textit{Stride = 4} is particularly informative. It suggests that although the model is still exposed to a broader temporal span, it no longer receives enough contiguous local structure to model next-item transitions as reliably as the denser sliding variants. In other words, long-range exposure alone is not sufficient if the sampling pattern removes too much of the short-range sequential signal. From a systems perspective, this means higher strides are best viewed as a last-resort efficiency knob: they are useful when memory or runtime constraints would otherwise make long-sequence training infeasible, but they should not be expected to preserve the full benefit of standard sliding-window training on datasets whose predictive signal is strongly order-sensitive.

\section{Conclusion and Implications}
We deliver a fully operational, open-source pipeline for long-sequence recommendation with sliding windows, augment it with ablations, and introduce \textit{k}-shift embeddings that unlock million-scale vocabularies on commodity GPUs. Since \cite{joshi2024sliding} did not specify model architecture details, we employed a generic transformer-based sequential recommendation model to ensure clarity and reproducibility. Our choice of datasets was influenced by their similarity to the Netflix data structure and feasibility. The result is a practical method that improves retrieval quality while making the accuracy–compute trade-offs explicit and tunable. Our experiments show that sliding-window exposure reliably boosts next-item ranking quality on public data, with transparent training-time overheads. As shown in Table \ref{tab:results}, the sliding window approach performs best in \textit{All-Sliding} mode (\textbf{RQ2}). Nevertheless, the original study \cite{joshi2024sliding} demonstrated that the \textit{Mixed-1000} mode yielded superior performance. Consequently, the optimal method is contingent upon the dataset and the maximum window size implemented in the ``control" section of the sliding window training. Strided variants map out controllable speed–quality trade-offs, cutting training time while preserving much of the long-context benefit. Moreover, \emph{k}-shift embeddings substantially reduce memory needs for large vocabularies with minimal loss, making long-sequence training feasible in low-resource environments.

The main implication of this paper is not that one sliding-window variant universally dominates, but that long-sequence exposure can be made practical, reproducible, and informative under resource constraints. On Retailrocket, \textit{All-Sliding} outperforms the mixed variants, whereas prior industrial results favored \textit{Mixed-1000}; this difference indicates that the best regime is dataset-dependent and likely mediated by factors such as sequence-length distribution, temporal drift, and interaction sparsity. Our stride ablation further shows that training-time savings are real but not free: moderate strides preserve much of the benefit, while larger strides can disrupt local sequential structure, as reflected by the rise in perplexity. Likewise, \emph{k}-shift embeddings are useful because they reduce memory pressure enough to make large-vocabulary training possible on modest hardware, but their effectiveness depends on keeping collision pressure within a tolerable range. Finally, because we focus on perplexity, MRR, and Recall rather than full-ranking metrics such as mAP, our conclusions should be interpreted primarily in the context of efficient next-item retrieval rather than exhaustive ranked-list quality. 

For broader impact, we have made our source code publicly available (\href{https://github.com/souradipp76/RecSysSWT}{https://github.com/souradipp76/RecSysSWT}) and demonstrated the methodology on public data, providing a transparent foundation for future work on low-resource long-sequence recommendation.\\

\bibliographystyle{unsrt}  
\bibliography{references}

@inproceedings{joshi2024sliding,
author = {Joshi, Swanand and Feng, Yesu and Hsiao, Ko-Jen and Zhang, Zhe and Lamkhede, Sudarshan},
title = {Sliding Window Training - Utilizing Historical Recommender Systems Data for Foundation Models},
year = {2024},
isbn = {9798400705052},
publisher = {Association for Computing Machinery},
address = {New York, NY, USA},
doi = {10.1145/3640457.3688051},
booktitle = {Proceedings of the 18th ACM Conference on Recommender Systems},
pages = {835–837},
numpages = {3},
location = {Bari, Italy},
series = {RecSys '24}
}

@article{zhang2023sockdef,
title = "SockDef: A Dynamically Adaptive Defense to a Novel Attack on Review Fraud Detection Engines",
author = "Youzhi Zhang and Sayak Chakrabarty and Rui Liu and Andrea Pugliese and Subrahmanian, \{V. S.\}",
note = "Publisher Copyright: IEEE",
year = "2024",
doi = "10.1109/TCSS.2023.3321345",
language = "English (US)",
volume = "11",
pages = "5253--5265",
journal = "IEEE Transactions on Computational Social Systems",
issn = "2329-924X",
publisher = "IEEE Systems, Man, and Cybernetics Society",
number = "4",
}

@inproceedings{bolonkin2024judicial,
  title={Judicial support tool: {F}inding the k most likely judicial worlds},
  author={Bolonkin, Maksim and Chakrabarty, Sayak and Molinaro, Cristian and Subrahmanian, VS},
  booktitle={International Conference on Scalable Uncertainty Management},
  pages={53--69},
  year={2024},
  organization={Springer}
}

@article{makarychev2023single,
  title={Single-pass pivot algorithm for correlation clustering. keep it simple!},
  author={Makarychev, Konstantin and Chakrabarty, Sayak},
  journal={Advances in Neural Information Processing Systems},
  volume={36},
  pages={6412--6421},
  year={2023}
}

@inproceedings{datta2023consistency,
author = {Datta, Arghya and Chakrabarty, Sayak},
title = {{On the consistency of maximum likelihood estimation of probabilistic principal component analysis}},
year = {2023},
publisher = {Curran Associates Inc.},
address = {Red Hook, NY, USA},
booktitle = {Proceedings of the 37th International Conference on Neural Information Processing Systems},
articleno = {1245},
numpages = {15},
location = {New Orleans, LA, USA},
series = {NIPS '23}
}

@article{chakrabarty2024free, 
doi = {10.21105/joss.07489}, 
year = {2025}, 
publisher = {The Open Journal}, 
volume = {10}, 
number = {108}, 
pages = {7489}, 
author = {Sayak Chakrabarty and Souradip Pal}, 
title = {Readme{R}eady: {F}ree and {C}ustomizable {C}ode {D}ocumentation with {LLM}s - {A} {F}ine-{T}uning {A}pproach}, 
journal = {Journal of Open Source Software} }

@article{Chakrabarty2025, 
doi = {10.21105/joss.07783}, 
year = {2025}, publisher = {The Open Journal}, 
volume = {10}, 
number = {108}, 
pages = {7783}, 
author = {Sayak Chakrabarty and Souradip Pal}, 
title = {{MM}-{P}o{E}: {M}ultiple {C}hoice {R}easoning via. {P}rocess of {E}limination using {M}ulti-{M}odal {M}odels}, 
journal = {Journal of Open Source Software} }

@article{li2023gpt4rec,
  title={{GPT4Rec}: A Generative Framework for Personalized Recommendation and User Interests Interpretation},
  author={Jinming Li and Wentao Zhang and Tiantian Wang and Guanglei Xiong and Alan Lu and G{\'e}rard G. Medioni},
  journal={ArXiv},
  year={2023},
  volume={abs/2304.03879},
  url={https://api.semanticscholar.org/CorpusID:258048480}
}

@inproceedings{10.1145/3357384.3357895,
author = {Sun, Fei and Liu, Jun and Wu, Jian and Pei, Changhua and Lin, Xiao and Ou, Wenwu and Jiang, Peng},
title = {BERT4Rec: {S}equential {R}ecommendation with {B}idirectional {E}ncoder {R}epresentations from {T}ransformer},
year = {2019},
isbn = {9781450369763},
publisher = {Association for Computing Machinery},
address = {New York, NY, USA},
doi = {10.1145/3357384.3357895},
booktitle = {Proceedings of the 28th ACM International Conference on Information and Knowledge Management},
pages = {1441–1450},
numpages = {10},
location = {Beijing, China},
series = {CIKM '19}
}

@inproceedings{zhai2024actions,
author = {Zhai, Jiaqi and Liao, Lucy and Liu, Xing and Wang, Yueming and Li, Rui and Cao, Xuan and Gao, Leon and Gong, Zhaojie and Gu, Fangda and He, Jiayuan and Lu, Yinghai and Shi, Yu},
title = {{Actions speak louder than words: trillion-parameter sequential transducers for generative recommendations}},
year = {2024},
publisher = {JMLR.org},
booktitle = {Proceedings of the 41st International Conference on Machine Learning},
articleno = {2414},
numpages = {26},
location = {Vienna, Austria},
series = {ICML'24}
}

@misc{ding2024longrope,
      title={Long{R}o{PE}: {E}xtending {LLM} {C}ontext {W}indow {B}eyond 2 {M}illion {T}okens}, 
      author={Yiran Ding and Li Lyna Zhang and Chengruidong Zhang and Yuanyuan Xu and Ning Shang and Jiahang Xu and Fan Yang and Mao Yang},
      year={2024},
      eprint={2402.13753},
      archivePrefix={arXiv},
      primaryClass={cs.CL}
}

@misc{zhu2024pose,
      title={{P}o{SE}: {E}fficient {C}ontext {W}indow {E}xtension of {LLM}s via {P}ositional {S}kip-wise {T}raining}, 
      author={Dawei Zhu and Nan Yang and Liang Wang and Yifan Song and Wenhao Wu and Furu Wei and Sujian Li},
      year={2024},
      eprint={2309.10400},
      archivePrefix={arXiv},
      primaryClass={cs.CL},
      url={https://arxiv.org/abs/2309.10400}, 
}

@inproceedings{geng2023recommendation,
author = {Geng, Shijie and Liu, Shuchang and Fu, Zuohui and Ge, Yingqiang and Zhang, Yongfeng},
title = {{Recommendation as Language Processing (RLP): A Unified Pretrain, Personalized Prompt \& Predict Paradigm (P5)}},
year = {2022},
isbn = {9781450392785},
publisher = {Association for Computing Machinery},
address = {New York, NY, USA},
doi = {10.1145/3523227.3546767},
booktitle = {Proceedings of the 16th ACM Conference on Recommender Systems},
pages = {299–315},
numpages = {17},
location = {Seattle, WA, USA},
series = {RecSys '22}
}

@inbook{Craswell2009,
  address = {Boston, MA},
  author = {Craswell, Nick},
  booktitle = {{Encyclopedia of Database Systems}},
  doi = {10.1007/978-0-387-39940-9_488},
  editor = {Liu, Ling and Özsu, M. Tamer},
  isbn = {978-0-387-39940-9},
  pages = {1703--1703},
  publisher = {Springer US},
  title = {Mean Reciprocal Rank},
  year = 2009
}

@misc{desai2022tradeoffsmodelsizelarge,
      title={{The trade-offs of model size in large recommendation models : {A} 10000 $\times$ compressed criteo-tb {DLRM} model (100 {GB} parameters to mere 10{MB})}}, 
      author={Aditya Desai and Anshumali Shrivastava},
      year={2022},
      eprint={2207.10731},
      archivePrefix={arXiv},
      primaryClass={cs.LG},
      url={https://arxiv.org/abs/2207.10731}, 
}

@article{DBLP:journals/corr/abs-1909-02107,
  author       = {Hao{-}Jun Michael Shi and
                  Dheevatsa Mudigere and
                  Maxim Naumov and
                  Jiyan Yang},
  title        = {{Compositional {E}mbeddings {U}sing {C}omplementary {P}artitions for {M}emory-{E}fficient {R}ecommendation {S}ystems}},
  journal      = {CoRR},
  volume       = {abs/1909.02107},
  year         = {2019},
  url          = {http://arxiv.org/abs/1909.02107},
  eprinttype    = {arXiv},
  eprint       = {1909.02107},
  bibsource    = {dblp computer science bibliography, https://dblp.org}
}

@inproceedings{10356156,
  author={Chengjie, Yang and Wei, Qi},
  booktitle={2023 IEEE International Conference on e-Business Engineering (ICEBE)}, 
  title={{Taobao {U}ser {P}urchase {B}ehavior {P}rediction {A}nd {F}eature {A}nalysis {B}ased On {E}nsemble {L}earning}}, 
  year={2023},
  volume={},
  number={},
  pages={205-209},
  doi={10.1109/ICEBE59045.2023.00027}
}

@article{chakrabarty2026pixrec,
  title={{PixRec: Leveraging Visual Context for Next-Item Prediction in Sequential Recommendation}},
  author={Chakrabarty, Sayak and Pal, Souradip},
  journal={arXiv preprint arXiv:2601.06458},
  year={2026}
}

@article{chakrabarty2025timeconstrainedrecsys,
  title={{Time-Constrained Recommendations: Reinforcement Learning Strategies for E-Commerce}},
  author={Chakrabarty, Sayak and Pal, Souradip},
  journal={arXiv preprint arXiv:2512.13726},
  year={2025}
}

@article{chakrabarty2025mm,
  title={{MM-PoE: Multiple Choice Reasoning via. Process of Elimination using Multi-Modal Models}},
  author={Chakrabarty, Sayak and Pal, Souradip},
  journal={Journal of Open Source Software},
  volume={10},
  number={108},
  pages={7783},
  year={2025}
}




\end{document}